\documentclass[10pt,twocolumn,letterpaper]{article}

\usepackage[pagenumbers]{cvpr} 

\usepackage[dvipsnames]{xcolor}

\usepackage[accsupp]{axessibility}

\definecolor{cvprblue}{rgb}{0.21,0.49,0.74}
\usepackage[pagebackref,breaklinks,colorlinks,citecolor=cvprblue]{hyperref}

\def\paperID{4430} 
\def\confName{CVPR}
\def\confYear{2024}

\title{GEARS: Local Geometry-aware Hand-object Interaction Synthesis}

\author{Keyang Zhou$^{1, 2}$ \quad Bharat Lal Bhatnagar$^{1, 2}$ \quad Jan Eric Lenssen$^{2}$ \quad Gerard Pons-Moll$^{1, 2}$\vspace{0.1cm} \\
 $^1$University of Tübingen, Germany \\
 $^2$Max Planck Institute for Informatics, Saarland Informatics Campus, Germany
}

\def\etal{\emph{et al}\onedot}

\renewcommand{\vec}[1]{\boldsymbol{#1}}
\newcommand{\mat}[1]{\mathbf{#1}}

\let\oldtwocolumn\twocolumn
\renewcommand\twocolumn[1][]{
	\oldtwocolumn[{#1}{
    	\begin{center}	\includegraphics[width=\linewidth]{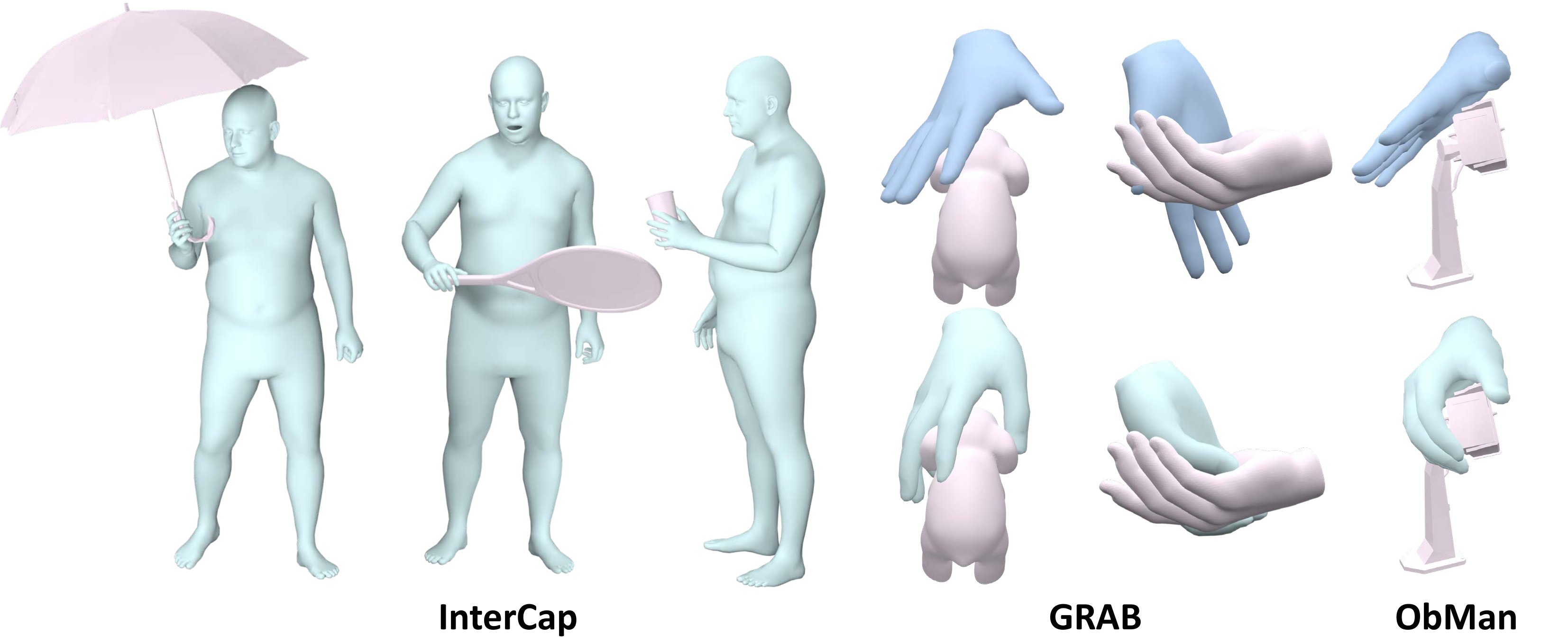}
            \captionof{figure}{We propose GEARS, a method to synthesize sequence of hand poses during interaction with an object. GEARS takes hand and object trajectory as input. It generates realistic hand poses that are well-adapted to object surface, irrespective of object category and size. We show sample results on different datasets. Hands colored in blue are inputs while hands colored in cyan are our predictions.}
            \label{fig:teaser}
		\end{center}
	}]
}

\begin{document}
\maketitle

\begin{abstract}
Generating realistic hand motion sequences in interaction with objects has gained increasing attention with the growing interest in digital humans. Prior work has illustrated the effectiveness of employing occupancy-based or distance-based virtual sensors to extract hand-object interaction features. Nonetheless, these methods show limited generalizability across object categories, shapes and sizes. We hypothesize that this is due to two reasons: 1) the limited expressiveness of employed virtual sensors, and 2) scarcity of available training data. To tackle this challenge, we introduce a novel joint-centered sensor designed to reason about local object geometry near potential interaction regions. The sensor queries for object surface points in the neighbourhood of each hand joint. As an important step towards mitigating the learning complexity, we transform the points from global frame to hand template frame and use a shared module to process sensor features of each individual joint. This is followed by a spatio-temporal transformer network aimed at capturing correlation among the joints in different dimensions. Moreover, we devise simple heuristic rules to augment the limited training sequences with vast static hand grasping samples. This leads to a broader spectrum of grasping types observed during training, in turn enhancing our model's generalization capability. We evaluate on two public datasets, GRAB and InterCap, where our method shows superiority over baselines both quantitatively and perceptually.
\end{abstract}

\begin{figure*}[t]
    \centering
    \includegraphics[width=\linewidth]{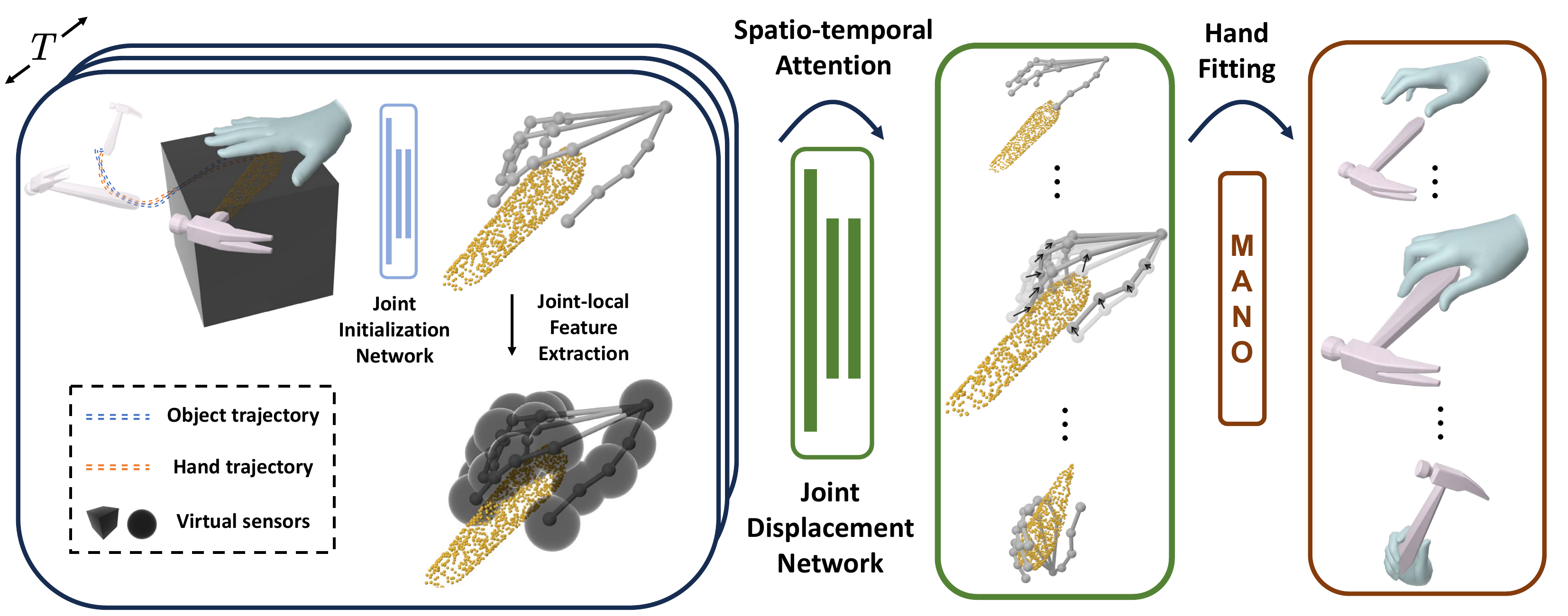}
    \caption{An overview of our method. The input consists of the hand trajectory, object trajectory and object template mesh. For each time frame, the object mesh is cropped with a cube-shaped virtual sensor positioned and oriented based on the wrist. The cropped object points together with the hand trajectory are fed to the Joint Initialization Network to predict coarse joints locations. We then place more fine-grained geometry sensors at each joint to extract joint-local object features. The features are subsequently processed by the Joint Displacement Network to refine the initialized joints. Finally, we fit MANO hand model~\cite{MANO:SIGGRAPHASIA:2017} to the joints to get the hand mesh sequence.}
    \label{fig:overview}
\end{figure*}

\section{Introduction}
\label{sec:intro}
We humans mostly rely on hands to interact with different objects in the surrounding environment. Learning the high-dimensional space of plausible hand-object interactions is an important and challenging task that needs to be solved in many applications. These include modeling digital humans in Augmented and Virtual Reality, or reasoning about potential grasps in robotics. 

Real world objects can largely differ in size, topology and geometry. Learning a model which can adapt to object surface is a particularly demanding task, especially when existing dynamic hand-object interaction data is very scarce. One crucial factor determining generalization capability lies in how the object is encoded relative to the hand. Previous work~\cite{zhang2021manipnet,taheri2023grip} proposed to use occupancy-based or distance-based virtual sensors to represent local surface geometry. However, these features have two limitations. First, they are inherently constrained by their expressiveness. Occupancy-based sensor attaches an occupancy grid to the hand. Occupancy grids with a low resolution can only detect coarse object geometry. On the other hand, increasing the grid resolution would result in an exponential increase in feature size. Distance-based sensor measures the distance from a fixed set of basis points rigidly attached to the hand to the closest points on the object surface. It gives more fine-grained features and it is also less computationally expensive. However, a discrete collection of hand-to-object distance cannot faithfully describe local object geometry properties such as normal directions and curvature. Moreover, features computed by both of the aforementioned sensors are global with respect to the hand, which means it is difficult to model the intricate correlation between the movement of each finger. As the result, these methods exhibit limited generalization capabilities to unseen objects of different sizes.

The ability of humans to perform dexterous object manipulations is attributed to the dense tactile sensory receptors in the skin. Thus, we hypothesize that the ability to reason about local geometry is key to generalization to arbitrary surfaces. Inspired by this, we propose a novel hand-object interaction sensor which is local to every hand joint. Specifically, we establish a canonical frame at each joint, and use a shared module to process local object points within a small radius of the joint. This way, the module learns joint-agnostic local features, which are highly generalizable from limited training data. We further fuse together features at each joint by self-attention operations, enabling the model to learn the compositional relationship between different joints in forming the hand pose.

Due to the limited availability of dynamic human-object interaction data, we present a simple yet effective method for generating dynamic hand sequences from static grasps. Static hand grasping data is easily accessible and exhibits a diverse range of object geometry and grasping type. With our data augmentation procedure, we can turn them into artifical grasping sequences. We show that adding them to our training dataset can further improve the results.

Our contributions are as follows:
\begin{itemize}
    \item We propose a learning-based method to synthesize diverse hand motion sequences interacting with objects. Though trained only on small hand-held objects, we show that our model naturally generalizes to objects of larger sizes (see Figure.~\ref{fig:teaser}).

    \item We introduce a novel hand-object interaction sensor, which detects local object surface geometry relative to hand joints. This is proven essential to our model's generalization capabilities.

    \item With a simple yet effective data augmentation trick, we are able to utilizing the vast amount of existing static hand grasp data to train our model.
    
    \item Our code and pre-trained model will be released to enhance further research in this direction.
\end{itemize}

\section{Related Work}
\label{sec:related}

\textbf{Static Grasp Synthesis.}
Synthesizing stable hand grasps given target objects has been extensively studied in computer graphics~\cite{kry2006interaction} and robotics~\cite{shimoga1996robot,sahbani2012overview}. Conventional analytical approaches assume a simplified contact model and solve a constrained optimization problem to satisfy the force-closure condition~\cite{nguyen1988constructing,bicchi1995closure,zheng2005coping,el20113d}. In contrast, data-driven approaches generate grasp hypotheses from human demonstrations or annotated 3D data, and rank them according to certain quality metrics~\cite{bohg2013data,li2007data}. Modern robotic grasping simulators usually combine the merits of both~\cite{miller2004graspit,leon2010opengrasp}. Recently, there has been an increasing interest in training neural network-based models for hand grasp generation~\cite{hasson19obman,taheri2020grab,corona2020ganhand,jiang2021hand,zhu2021toward,brahmbhatt2020contactpose,karunratanakul2021skeleton,liu2023contactgen}. For example, \cite{karunratanakul2020grasping,jiang2021synergies} modeled the hand-object proximity as an implicit function. 

\textbf{Dynamic Grasp Synthesis.}
In comparison with static grasp synthesis, generating dynamic manipulation of objects is more challenging since it additionally requires dynamic hand and object interaction to be modeled. This task is usually approached by optimizing hand poses to satisfy a range of contact force constraints~\cite{liu2009dextrous,ye2012synthesis,mordatch2012contact,zhao2013robust}. With the advent of deep reinforcement learning, a number of work explored training hand grasping control policies in physics simulation~\cite{christen2022dgrasp,xu2023unidexgrasp,zhang2024artigrasp}. Hand motions generated by these works are physically plausible but lack natural variations. Zheng~\etal~\cite{Zheng_2023_CVPR} modeled hand poses in a canonicalized object-centric space, achieving category-level generalization for both rigid and articulated objects. More similar to our work are  ManipNet~\cite{zhang2021manipnet} and GRIP~\cite{taheri2023grip}, which utilized occupancy-based and distance-based sensors to extract local object features near the hand and then directly regressed hand poses from the features. We argue that these features are limited by resolution and they are global with respect to the hand, hence hindering generalization capability. In contrast, we adopt a novel joint-centered point-based sensor which captures local object geometry in finer details while enabling modeling the correlation among hand joints.

\textbf{Full-body Human-object Interaction Synthesis.} Generating realistic human motion sequences in 3D scenes has received considerable attention in recent years~\cite{wang2021synthesizing,hassan2021stochastic,zhao2022compositional,wang2022towards,zhang2022couch,jiang2023full,mir23origin}. However, these work usually models coarse body motion only and ignore fine-grained finger articulations. Another line of work focused on generating full-body motion for grasping~\cite{ghosh2023imos}. A typical solution to this problem is first generating the final static grasping pose and then using a motion infilling network to generate the intermediate poses~\cite{taheri2021goal,wu2022saga}. \cite{tendulkar2023flex} and~\cite{Zheng_2023_ICCV} leveraged the existing body pose prior and hand-only grasping prior to circumvent the limited diversity in available full-body grasping data. Braun~\etal~\cite{braun2023physically} adopted a physics-based approach, training separate low-level policies for the body and fingers, and then integrating them with a high-level policy which operates in latent space.

\textbf{Grasp Refinement.}
As consumer-level hand tracking devices including RGB/depth cameras, data gloves and IMUs become more accessible, it is relatively easy to acquire hands that are approximately correct but may contain noise and artifacts. Refining hand poses in accordance with hand-object interaction emerges as a practical research problem~\cite{taheri2020grab}. ~\cite{yang2021cpf} and~\cite{grady2021contactopt} proposed to identify the potential contact area on the object surface and subsequently adjust the hand to align with the predicted contact points. Limited by their contact representations, they can only handle static hand grasp. Zhou~\etal~\cite{zhou2022toch} improved upon them and extended the binary contact map representation to a spatio-temporal correspondence map, enabling the refinement of a hand motion sequence. We deviate from these work in our assumptions, as we only require hand and object trajectories as input.

\section{Method}
\label{sec:method}

Given the trajectories of a hand and an object in interaction, we aim to generate hand poses that align with the object motion. The object shape is assumed to be known. We tackle this problem in three steps. First, we estimate a coarse initial hand pose for each frame individually. We place virtual sensors on the initialized hand joints, detecting nearby object surface points and extracting hand-object interaction features based on these points. Local to each joint, the features are fed to a spatio-temporal attention network, which learns the correlation among hand joints and generates displacements to the initialized hand joints. Lastly, we solve an optimization problem and fit a parametric hand model to the predicted joints. See Figure.~\ref{fig:overview} for an overview of our method.

Specifically, our input consists of the hand trajectory $\left\{ \vec{w}^{t},\vec{R}_{H}^{t} \right\} _{t=1}^{T}$, the object trajectory $\left\{\vec{o}^{t},\vec{R}_{O}^{t} \right\} _{t=1}^{T}$ and the object template mesh $\vec{M}_O = \left\{ \vec{V}_O, \vec{F}_O\right\}$, with $\vec{w}^{t}, \vec{o}^t \in \mathbb{R}^3$ denoting hand and object translations and $\vec{R}_{H}^{t}, \vec{R}_{O}^{t} \in \text{SO}^3$ denoting hand and object global orientations respectively. We use the MANO model~\cite{MANO:SIGGRAPHASIA:2017} as our hand representation, which is parameterized by shape $\vec{\beta}$ and pose $\vec{\theta}$. Hence the hand trajectory is composed of the wrist joint coordinates and the global orientation of the target MANO hand at each frame.

\subsection{Joint Initialization Network}
Given the object position and orientation at frame $t$, we first obtain the object mesh at that frame by $\vec{V}_O^t = \vec{R}_{O}^{t}\vec{V}_O + \vec{o}^{t}$. In order to predict an initial hand pose, only the part of the object which is close to the wrist matters. Hence we crop the object by a cube-shaped virtual sensor $\vec{S}^t$ rigidly attached to the wrist. The resulting partial object mesh is denoted by ${\vec{M}_{O}^t}^{\prime} = \left\{ {\vec{V}_{O}^t}^{\prime}, {\vec{F}_O^t}^{\prime}\right\}$, where ${\vec{V}_{O}^t}^{\prime} = \{\vec{v}_i \in \vec{V}_O^t : \vec{v}_i \notin \vec{S}^t \} $ and ${\vec{F}_O^t}^{\prime} \subseteq \vec{F}_O^t$.

Let $\vec{P}^t \in \mathbb{R}^{N \times 3}$ denote the point cloud sampled on ${\vec{M}_{O}^t}^{\prime}$. We subsequently express the point cloud relative to the wrist:

\begin{align}
\tilde{\vec{P}}^t = {\vec{R}_{H}^{t}}^T \left( \vec{P}^t -  \vec{w}^{t}\right) .
\label{eq: cano}
\end{align}

We additionally sample the hand trajectory centered on the current frame. We sample $k$ frames both in the past and in the future, and express them relative to the wrist in a similar fashion as~\ref{eq: cano}. The inputs to the hand pose initialization module are
$\left[ \tilde{\vec{w}}^{t-k:t+k},\tilde{\vec{R}}_{H}^{t-k:t+k}, \tilde{\vec{P}}^t\right]$, where $\tilde{\vec{w}}^{t-k:t+k}$ and $\tilde{\vec{R}}_{H}^{t-k:t+k}$ are canonicalized sampled wrist positions and orientations respectively. In particular, we first use PointNet to extract a global feature vector from the partial point cloud $\vec{P}$. This feature vector is then concatenated with the trajectory and fed to a three-layer fully-connected network. The output of the network is denoted by $\vec{j}_\text{init}^t$, which represents the initialized coordinates relative to the wrist. The training loss for this module is defined by
\begin{align}
    L_\text{init} = \left\| \vec{j}_\text{init}-\vec{j}_\text{gt} \right\| _{2}^{2},
\end{align}
where $\vec{j}_\text{gt}$ denotes groundtruth joint coordinates.

\subsection{Local Geometry Sensor}
\begin{figure}[t]
    \centering
    \includegraphics[width=\linewidth]{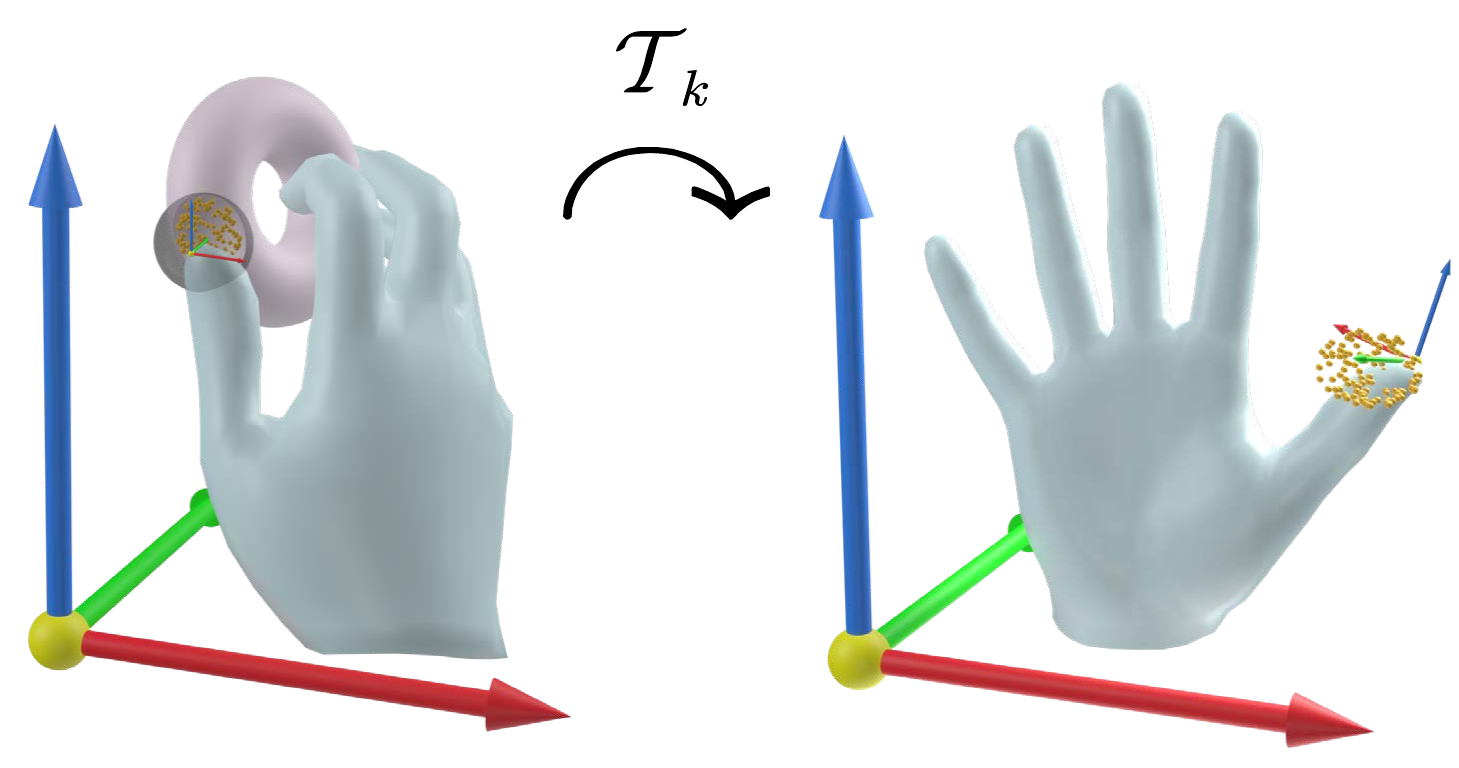}
    \caption{Visualization of our joint-local geometry sensor. (Left) Given the joints positions and the object mesh, we sample points on the object surface within a specified radius centered at each joint. The object points are represented in a joint-local frame. (Right) We transform the sampled object points from global frame to the canonical frame defined by the MANO template hand.}
    \label{fig:sensor}
\end{figure}
Although coarse and inaccurate, the initialized joints offer an indication of where the hand could potentially interact with the object. To refine the initial joint positions, we need to sense local geometry properties of the object near the interaction regions. 

We introduce a novel joint-centered point-based local geometry sensor to overcome these limitations. Specifically, given the predicted joints $\{\vec{j}_i\}_{i=1}^J$, we can utilize inverse kinematics to analytically derive the joint rotations which satisfy:

\begin{align}
\vec{j}_k - \vec{j}_{\text{pa}(k)} = \vec{R}_{k, \text{pa}(k)}\left(\Bar{\vec{j}}_k - \Bar{\vec{j}}_{\text{pa}(k)}\right),
\end{align}
where $\vec{R}_{k, \text{pa}(k)}$ is the relative angle between the $k$-th joint and its parent, and $\{\Bar{\vec{j}}_k\}$ are joints in the rest pose. In this manner, we can define the template frame of the $k$-th joint by

\begin{align}
\mathcal{T}_k = \prod_{i\in \text{A}\left( k \right)}{
\begin{array}{c|c}
\vec{R}_{i, \text{pa}(i)} & \vec{j}_i \\
\hline
\vec{1} & \vec{0}
\end{array}
},
\end{align}
where $\text{A}(k)$ denotes the list of ancestors of joint $k$ and $\mathcal{T}_k$ is the transformation which brings joint $k$ from the template frame to the global frame. 

By sampling object surface points within a given radius $r$ of the $k$-th joint along with their normal vectors, we get $\vec{F}_k = \{ \vec{P}_k, \vec{N}_k\}$, where $\vec{P}_k = \{ \vec{v}_i \in \vec{V}: \left\| \vec{v}_i - \vec{j}_k \right\| _{2}^{2} < r\}$. We then transform the sampled points to the template frame, by
\begin{align}
\Bar{\vec{F}}_k &= \{ \Bar{\vec{P}}_k, \Bar{\vec{N}}_k\} \\
&= \{\mathcal{T}_k^{-1}(\vec{P}_k - \vec{j}_k), \mathcal{T}_k^{-1}\vec{N}_k\}.
\end{align}

Since we now have the sampled object points in a joint-centered canonical frame, we apply a learnable module $f_\text{feat}$ to process the transformed points. Note that this module is shared between joints, which greatly reduces the learning complexity. We hence arrive at a hand-object interaction feature $\vec{f}_k = f_\text{feat}(\Bar{\vec{F}}_k)$ for each joint $k$. We implement $f_\text{feat}$ with a three-layer PointNet architecture.

\subsection{Joint Displacement Network}
With the local object features aggregated at each joint, we propose to use a transformer architecture to predict displacement vectors to the initialized joints. Achieving a visually plausible and smooth hand sequence requires modeling spatio-temporal inter-joint dependencies. Hence we apply the self-attention operation in both spatial and temporal dimensions. Concretely, we first project the initialized joint coordinates to per-joint embedding vectors with a fully-connected network $g_\text{embed}$:
\begin{align}
\vec{e}_k = g_\text{embed}(\mathcal{T}_k^{-1}\vec{j}_k),
\end{align}
where $k$ is the joint index. We concatenate the joint-local sensor features with the joint embeddings and obtain $\vec{X} = \text{concat}(\vec{f}, \vec{e})$, which is the input feature tensor for our transformer. Note that $X$ contains features for all the joints in all time frames. We divide $\vec{X}$ along its spatial and temporal dimension, and apply a self-attention function to them separately.

\begin{figure}[t]
    \centering
    \includegraphics[width=\linewidth]{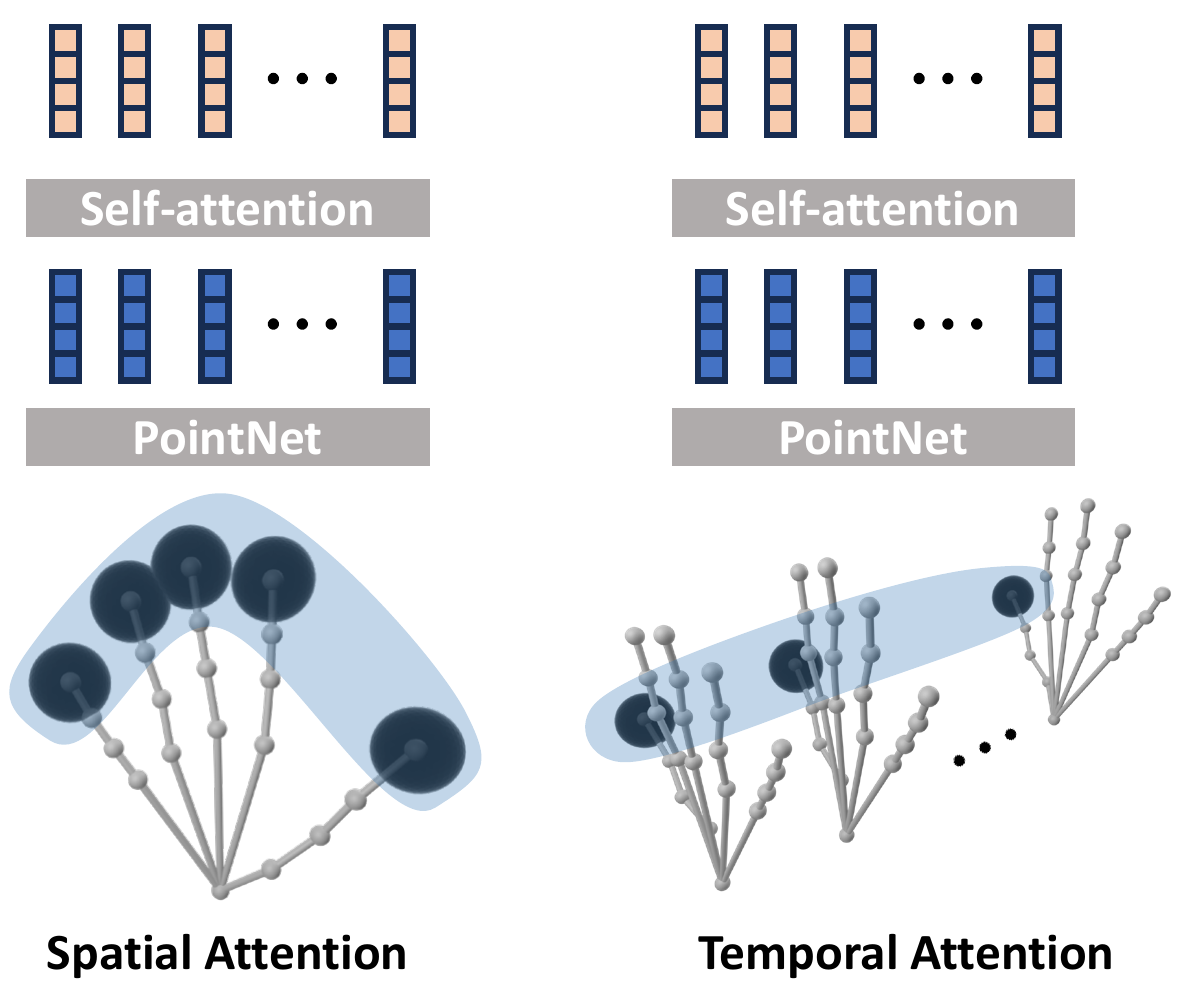}
    \caption{An illustration of spatial and temporal attention networks. We first process the features of each joint by PointNet. For spatial attention, every joint attends to every other joint of the same hand. While for temporal attention, a joint in one frame attends to the same joint in every other frame.}
    \label{fig:arch}
\end{figure}

\textbf{Spatial self-attention.} The spatial self-attention module divides $\vec{X}$ into batches of frames, and each batch contains joint features in a single frame, denoted by $\vec{X}_S$. This module takes the hands in different frames as static identities, and focuses on learning the correlations between different fingers. Following conventional self-attention operations, we linearly project $\vec{X}_S$ to queries $\vec{Q}_S$, keys $\vec{K}_S$ and values $\vec{V}_S$. The output feature is hence obtained by
\begin{align}
    \tilde{\vec{X}}_S &= \text{sa}(\vec{Q}_S,\vec{K}_S,\vec{V}_S) \\
    &= \text{softmax}\left(\frac{\vec{Q}_S \vec{K}_{S}^{T}}{\sqrt{l}}\right)\vec{V}_S,
\end{align}
where $l$ is the length of key, query and value vectors. See Fig.~\ref{fig:arch} (Left) for an illustration.

\textbf{Temporal self-attention.} On the other hand, the temporal self-attention module divides $\vec{X}$ into batches of joints, and each batch contains features of a specific joint across the whole sequence, denoted by $\vec{X}_T$. This module models the trajectory of each individual joint, ensuring that all joints move in a temporally smooth and consistent manner. We similarly project $\vec{X}_T$ to queries $\vec{K}_T$, keys $\vec{Q}_T$ and values $\vec{V}_T$ respectively. The module output is
\begin{align}
    \tilde{\vec{X}}_T &= \text{sa}(\vec{Q}_T,\vec{K}_T,\vec{V}_T).
\end{align}
See Fig.~\ref{fig:arch} (Right) for an illustration.

The Joint Displacement Network consists of interleaving spatial and temporal self-attention modules. The output of the last module is fed to a linear layer to produce the joint displacement vectors $\bar{\vec{d}}$ in template frame. As the last step, we utilize the pose transformation derived from IK previously to transform $\bar{\vec{d}}$ back to global frame:
\begin{align}
\vec{d}_k = \left(\prod_{i\in A(k)}{\vec{R}_{i, \text{pa}(i)}}\right) \bar{\vec{d}_k}.
\end{align}
The training loss for this module is defined by
\begin{align}
    L_\text{disp} = \left\| \vec{j}_\text{disp} + \vec{d} -\vec{j}_\text{gt} \right\| _{2}^{2}.
\end{align}

\subsection{Hand Fitting}

With the predicted sequence of hand joints $\vec{j}$, we need to recover the hand meshes. This is done by minimizing

\begin{align}
    \mathcal{L}(\vec{\beta}, \vec{\theta}) = \left\| \mathcal{J}\left( H\left( \vec{\beta} ,\vec{\theta} \right) \right) - \vec{j} \right\| _{2}^{2}+ \mathcal{L}_\text{reg}(\vec{\beta}, \vec{\theta}) \textnormal{,}
    \label{eq:fitting}
\end{align}
where $\mathcal{J}$ is the function which takes hand vertices as input and outputs joint coordinates. The second term of (\ref{eq:fitting}) regularizes the shape and pose parameters of MANO,
\begin{multline}
     \mathcal{L}_\text{reg}(\vec{\beta}, \vec{\theta}) = w_1\left\| \vec{\beta} \right\| ^2+w_2\sum_{t=1}^T{\left\| \vec{\theta}^t \right\| ^2} \\ + w_3\sum_{t=1}^{T-1}{\left\| \vec{\theta}^{t+1} - \vec{\theta}^{t} \right\| ^2} 
     + w_4\sum_{t=2}^{T-1}{\sum_{i=1}^J{\left\| \ddot{\vec{j}}_i^t\right\|} },
\end{multline}
where we enforce temporal smoothness by regularizing both the first and the second time derivatives of the joints. 

\subsection{Data Synthesis}
Accurately capturing hand motion sequences, especially in presence of interacting objects, is a particularly challenging task. Sophisticated solutions usually involve expensive marker-based MoCap systems. As a result, there are only few dynamic hand-object interaction datasets available for use. Nevertheless, capturing the hand in a static pose while grasping an object is relatively straightforward. We can have a much larger training set if we are able to utilizing the widely available static hand grasping datasets. In the following, we introduce a simple yet efficient way to synthesize hand sequences from static poses.

Given mesh of a static hand grasping an object, we first fit MANO model to the hand to get the target joint rotations $\vec{P}^T$, global orientation $\mat{R}^T$ and translation $\vec{d}^T$, where $T$ is the desired sequence length. We then generate a source hand as the first frame of the sequence, where the pose is generated by adding a small random Gaussian noise to the mean MANO pose. Similarly, we perturb $\mat{R}^T$ with Gaussian noise to get the global orientation of the initial hand. Next, we compute the average distance moved by the hand per frame from GRAB. The initial translation is determined by moving along the negative normal direction of the target hand palm by this distance.

\begin{figure}[t]
    \centering
    \includegraphics[width=\linewidth]{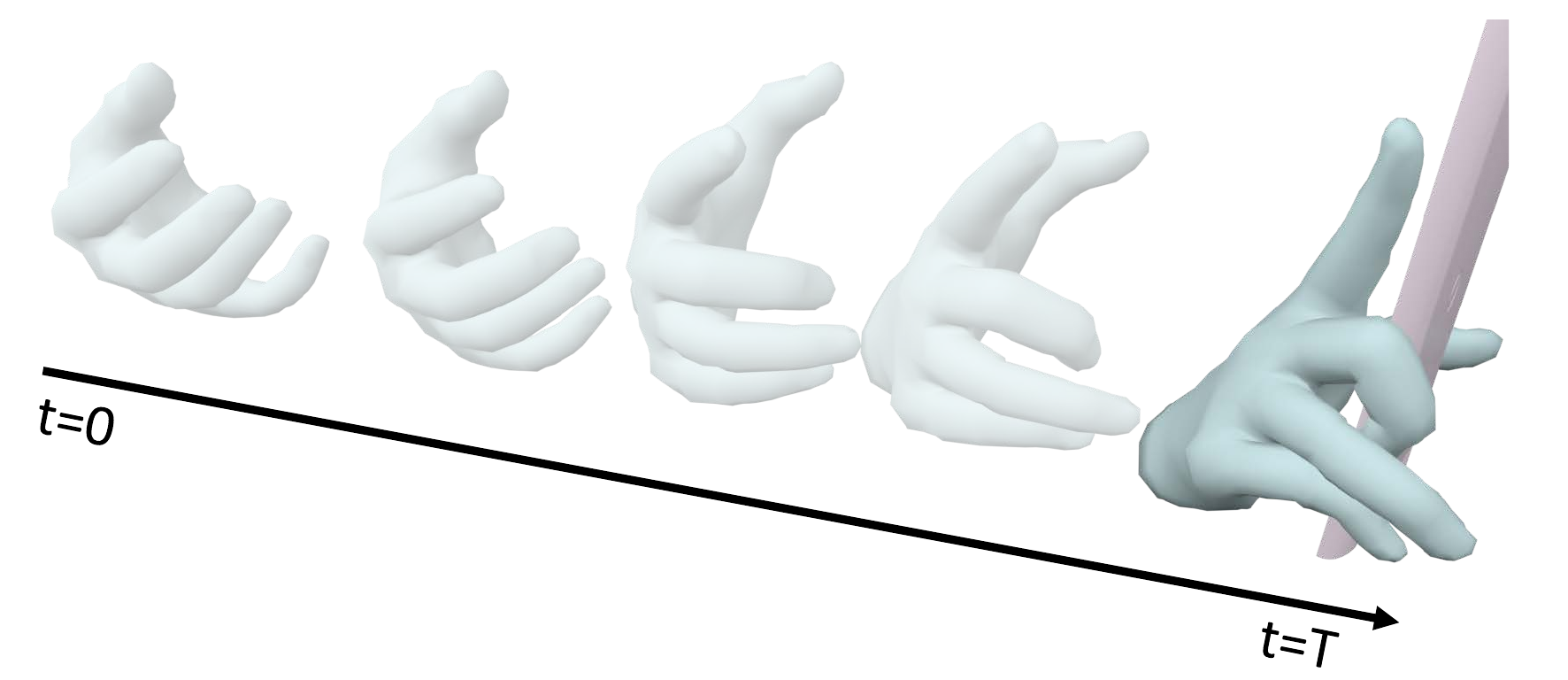}
    \caption{A sample training sequence synthesized by our heuristic rule. At the rightmost side of the time axis is a static grasping pose from ObMan~\cite{hasson19obman}. We synthesize intermediate poses by interpolating joint angles from the mean MANO pose.}
    \label{fig:aug}
\end{figure}

To obtain hand meshes in intermediate time steps, we apply linear interpolation to hand translation and spherical linear interpolation to joint rotations:
\begin{align}
\vec{d}^t &= (1-t)  \vec{d}^0 + t  \vec{d}^T \\
\vec{P}^t &= \text{SLERP}(\vec{P}^0, \vec{P}^T, t) \\
\vec{R}^t &= \text{SLERP}(\vec{R}^0, \vec{R}^T, t).
\end{align}
Generating sequences in this way could result in hand-object intersections. Rather than relying on path planning algorithms to prevent collisions, we simply compute the highest intersection volume of a sequence and eliminate sequences with intersection volume surpassing a predefined threshold. See Figure.~\ref{fig:aug} for a sample sequence.

\subsection{Implementation Details}
For Joint Initialization Network, the side length of the cube sensor is 18cm. We sample 2000 points on the partial object mesh as input to PointNet. We uniformly sample 10 frames in the past and in the future within a 1 second time window to compute the trajectory feature. When querying for joint-local object points, we use the sphere sensor with a radius of 2.5cm. A maximum of 300 points are sampled in the neighbourhood of each joint.

\section{Experiments}
\label{sec:experiments}

\begin{table*}
    \centering
    \begin{tabular}{l | cccc}
         & MPJPE (mm) $\downarrow$ &  PD (mm) $\downarrow$&  IV ($\text{cm}^3$) $\downarrow$& C-IoU (\%) $\uparrow$\\
        \hline
        TOCH &  8.18&  5.37&  2.72& 20.1\\
        \hline
        ManipNet &  9.32&  5.66&  3.21& 18.3\\
        \hline
        GRIP &  7.71&  4.80&  2.51& 19.9\\
        \hline
        GEARS (ours) &  \textbf{7.24}&  \textbf{4.36}&  \textbf{2.24}& \textbf{22.7}\\
         
    \end{tabular}
    \caption{We quantitatively compare GEARS to other baselines on the GRAB dataset. Each model is trained with the same amount data, including the synthetic sequences generated from ObMan.}
    \label{tab:grab}
\end{table*}

\begin{table}
    \centering
    \begin{tabular}{l | cc}
         & PD (mm) $\downarrow$&  IV ($\text{cm}^3$) $\downarrow$\\
        \hline
        ManipNet &  8.22& 6.15\\
        \hline
        GRIP &  7.92& 5.68\\
        \hline
        GEARS (ours) &  \textbf{7.44}& \textbf{5.21}\\
         
    \end{tabular}
    \caption{Quantitative comparison on InterCap. We evaluate on a selected subset of objects where hand interaction is involved.}
    \label{tab:intercap}
\end{table}

\begin{table}
    \centering
    \begin{tabular}{l | ccc}
         & MPJPE (mm) $\downarrow$ &  C-IoU (\%) $\uparrow$\\
         \hline
         $r=0$ &   9.34& 14.8\\
         $r=0.02$ &  7.28 & 22.1\\
         $r=0.03$ &  7.24 & 21.9\\
         \hline
        w/o displacement &  9.63 & 13.2\\
        w/o attention &  7.85 & 18.4\\
        w/o synthetic &  7.31 & 20.6\\
        \hline
        Ours (iterative) &  7.37 & 19.2\\
        \hline 
        Ours (full)&  \textbf{7.24} & \textbf{22.7}\\
         
    \end{tabular}
    \caption{Ablaion studies evaluated on GRAB. The variable $r$ refers to the radius of joint-local sensor in millimeters.}
    \label{tab:ablation}
\end{table}


\subsection{Datasets}
\label{sec:exp_datasets}
\noindent
\textbf{GRAB}. We train GEARS on GRAB~\cite{taheri2020grab}, a large-scale MoCap dataset for whole-body grasping. GRAB contains interaction sequences with $51$ objects. Following the official protocol, we select $10$ objects for validation and testing, and train with the rest. Due to symmetry of the two hands, we flip left hands to increase the amount of training data. We further augment the training set by transferring grasps to objects of varying sizes, following~\cite{zhou2022toch}.

\noindent
\textbf{InterCap}. InterCap~\cite{huang2022intercap} is a dataset of whole-body human-scene interaction captured by multiview RGB-D cameras. It features frame-wise pseudo-groundtruth annotations for body, hand and 6D object poses, which are reconstructed by jointly reasoning about human and object contact areas. As we solely focus on hand-object interaction, we consider a subset of objects where the hand is in interaction.

\noindent
\textbf{ObMan}. ObMan~\cite{hasson19obman} is a static hand grasping dataset. It consists of object models taken from Shapenet~\cite{chang2015shapenet} and synthetic hand grasps generated by the robotic grasping software GraspIt~\cite{miller2004graspit}. Since it only has static hand poses, we cannot directly train on it. Instead, we apply the data synthesis technique and generate 200 sequences for training and testing. Each sequence has a fixed length of 60 frames.

\subsection{Metrics}
\label{sec:exp_metrics}
\noindent
\textbf{Mean Per-Joint Position Error (MPJPE)}. We report the average Euclidean distance between predicted and groundtruth 3D hand joints.

\noindent
\textbf{Penetration Depth (PD)}. Penetration depth is the minimum distance required for moving a mesh to make it no longer in intersection with another mesh. We approximate it by finding the maximum vertex-to-object distance for all the penetrating hand vertices.

\noindent
\textbf{Intersection Volume (IV)}. We measure hand-object inter-penetration by voxelizing hand and object meshes and reporting the volume of voxels occupied by both. However, interpreting this metric in isolation could be misleading, since it does not account for non-effective grasping artifacts.

\noindent
\textbf{Contact IoU (C-IoU)}. This metric evaluates the Intersection-over-Union between the groundtruth binary hand-object contact map and the contact map of predicted hands. The contact map is defined on the object surface. It takes a value of 1 if a hand vertex is within $\pm 2 \mathrm{mm}$ of an object vertex and 0 otherwise.

\subsection{Baselines}
\label{sec:exp_baselines}
\noindent
\textbf{TOCH}~\cite{zhou2022toch} is an object-centric model designed for refining noisy hand-object interaction sequences. We tailor it to our task by feeding it with the groundtruth hand trajectory and replacing the noisy hands in the training set with flat hands.

\noindent
\textbf{ManipNet}~\cite{zhang2021manipnet} relies on both occupancy-based and distance-based sensors to generate dexterous hand motions. Since the original work assumed a different hand model, we adapt it to MANO to compare on a fair ground. 

\noindent
\textbf{GRIP}~\cite{taheri2023grip} takes body arm trajectory as input to generate hand poses. It employs a standalone module to denoise the arm trajectory and obtain hand trajectory. For fair comparison, we directly provide GRIP with input hand trajectories.

\subsection{Quantitative and Qualitative Evaluation}
\label{sec:exp_quantitative}
\begin{figure*}[t]
    \centering
    \includegraphics[width=\linewidth]{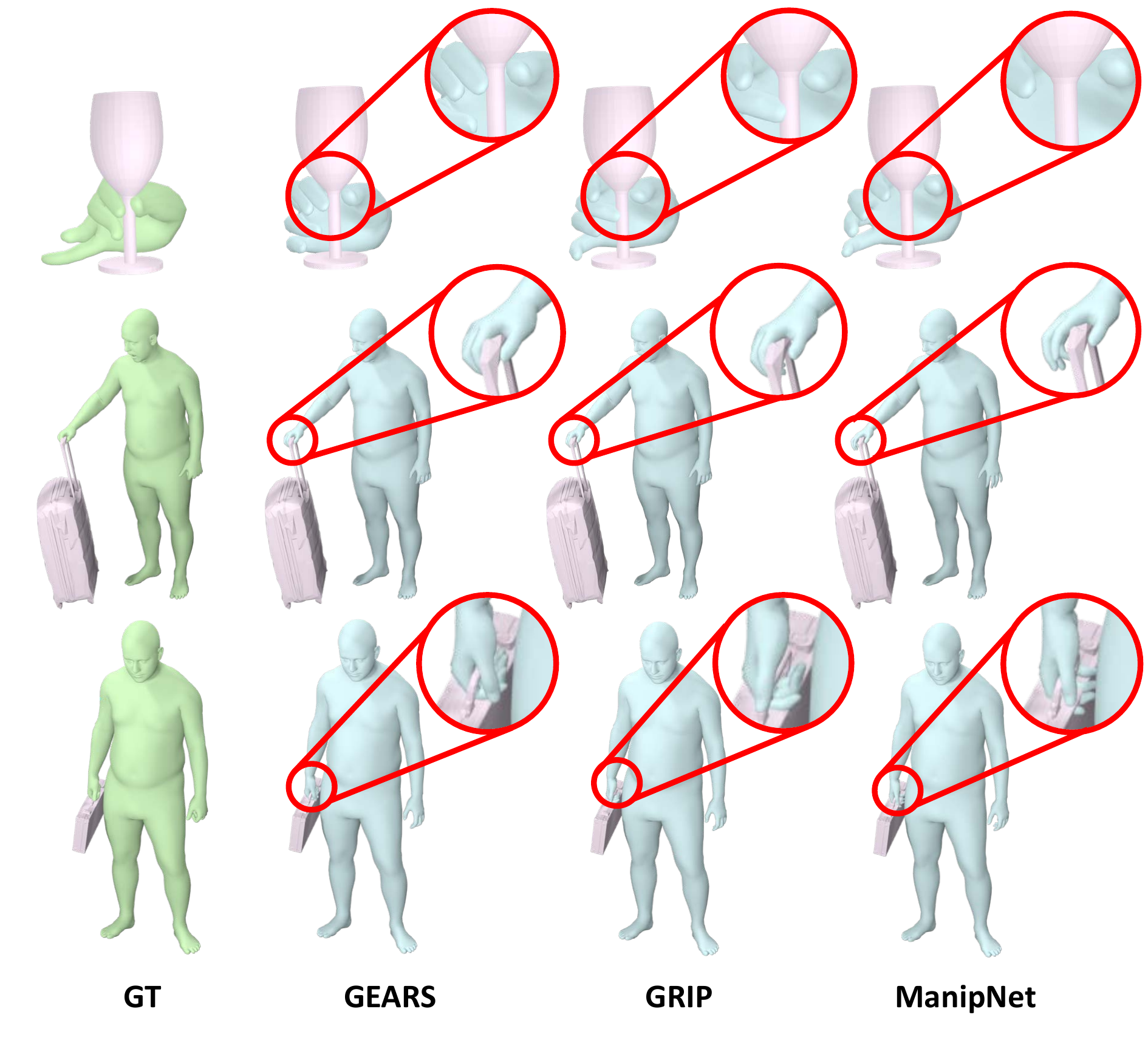}
    \caption{Qualitative results on GRAB (top row) and InterCap (bottom two rows). GEARS makes effective contact with the objects while avoiding hand-object inter-penetration.}
    \label{fig:qualitative}
\end{figure*}

To verify that our GEARS generates realistic interaction sequences, we first evaluate our method on GRAB and compare with the aforementioned baselines. The results are reported in Table.~\ref{tab:grab}. GEARS outperforms other baselines on all four metrics, which clearly demonstrates the advantage of our method. We can observe that although both ManipNet and GRIP rely on distance-based sensors, GRIP achieves better performance both in terms of joint accuracy and inter-penetration score. We hypothesize that it could attribute to the two-stage approach followed by GRIP. Similar to us, GRIP generates a coarse hand first and subsequently refined it. Moreover, TOCH incurs a higher MPJPE but also achieves higher contact IoU than GRIP. This observation shows that a higher joint error doesn't necessarily indicate worse grasping quality. TOCH leverages an object-centric interaction representation, which naturally encourages hand-object contact. See Figure.~\ref{fig:qualitative} (top row) for qualitative results on GRAB.

GRAB contains mostly small-to-medium sized household objects. To assess our model's generalization capability to larger objects, we evaluate on the InterCap dataset. We exclude TOCH from this comparison because the object-centric contact map used by TOCH is highly sensitive to object size. Since the groundtruth hand pose annotations of InterCap are not accurate enough, we only report penetration depth and intersection volume, see Table.~\ref{tab:intercap}. Compared to GRIP and ManipNet, GEARS incurs less penetration with the objects. Note that all three methods report higher numbers than on GRAB. It can be partially explained by the fact that the input hand trajectory provided by InterCap may exhibit a certain degree of noise. See Figure.~\ref{fig:qualitative} (bottom rows) for a qualitative comparison on InterCap.

\subsection{Ablation Studies}
We ablate different components of GEARS and report the change in performance on GRAB to further justify our proposed method, see Table~\ref{tab:ablation}. We first evaluate how sensitive is the model to different sensor radius. Zero radius means that sensor features are neglected by the network. We can observe that as long as the radius is set within a reasonable range, it doesn't have a significant impact on performance.

Moreover, we train three baseline models, for which i) the Joint Displacement Network is removed; ii) spatio-temporal attention is replaced by fully-connected layers; iii) additional synthetic training sequences are not used. It's clear that the Joint Displacement Network plays the most important role in our architecture. This agrees with our intuition that local object geometry features are essential to fine-grained placement of joints.

Lastly, we design an iterative baseline, where at inference time the output of the Joint Displacement Network is fed back to itself as input. We expect that one more round of pose refining would further improve the generation quality. Surprisingly, the iterative refining approach doesn't bring any benefit. We hypothesize that the underlying reason could be distributional shift of test data, since the Joint Displacement Network has only seen the output of Joint Initialization Network during training. 

\label{sec:exp_ablation}

\section{Conclusion}
\label{sec:conclusion}
We present GEARS, a learning-based method for generating hand interaction sequences given hand and object trajectories. The main insight which makes GEARS effective is the novel joint-centered point-based sensor which captures local geometry properties of the target object. Furthermore, we design a spatio-temporal self-attention architecture to process joint-local features and learn the correlation among hand joints during interaction. GEARS is capable of generalizing across objects of varying sizes and categories. We show that GEARS outperforms previous methods in terms of generation quality and generalizability.

\noindent
\textbf{Acknowledgements} 
This work is supported by the German Federal Ministry of Education and Research (BMBF): Tübingen AI Center, FKZ: 01IS18039A. This work is funded by the Deutsche Forschungsgemeinschaft (DFG, German Research Foundation) - 409792180 (Emmy Noether Programme, project: Real Virtual Humans). Gerard Pons-Moll is a member of the Machine Learning Cluster of Excellence, EXC number 2064/1 – Project number 390727645. The project was made possible by funding from the Carl Zeiss Foundation.

{
    \small
    \bibliographystyle{ieeenat_fullname}
    \bibliography{main}
}


\end{document}